\documentclass[letterpaper, 10 pt,journal, twoside]{IEEEtran}  

\IEEEoverridecommandlockouts                              

\usepackage{graphics} 
\usepackage{epsfig} 
\usepackage{times} 
\usepackage{amsmath} 
\usepackage{amssymb}  
\usepackage{amsfonts}
\usepackage[colorlinks,linkcolor=blue,anchorcolor=blue,citecolor=blue]{hyperref}
\usepackage[table]{xcolor}
\usepackage{hyperref}
\usepackage{booktabs}
\usepackage{multirow}
\usepackage{graphicx}
\usepackage{subfigure}
\usepackage{cite}
\usepackage{float}
\usepackage{makecell}
\usepackage{algorithm}
\usepackage[noend]{algorithmic}

\usepackage{etoolbox}

\hypersetup{
    colorlinks=true,
    citecolor=blue    
}

\makeatletter
\patchcmd{\@makecaption}
  {\scshape}
  {}
  {}
  {}
\makeatletter
\patchcmd{\@makecaption}
  {\\}
  {.\ }
  {}
  {}
\makeatother
\def\tablename{Table}

\graphicspath{{img/}}

\title{Multi-Agent Trajectory Prediction with Difficulty-Guided Feature Enhancement Network}

\author{Guipeng Xin$^{1}$, Duanfeng Chu$^{1,\dag}$, Liping Lu$^{2}$, Zejian Deng$^{3}$, Yuang Lu$^{1}$, and Xigang Wu$^{1}$
 
\thanks{This work is supported in part by the National Key R\&D Program of China (2021YFB2501104), the Natural Science Foundation of Hubei Province for Distinguished Young Scholars (2022CFA091), and Wuhan Science and Technology Major Project (2022013702025184).({\it Corresponding author: Duanfeng Chu.})}

\thanks{$^{1}$Guipeng Xin, Duanfeng Chu, Yuang Lu and Xigang Wu are with the Intelligent Transportation Systems Research Center, Wuhan University of Technology, Wuhan 430063, China (e-mail: xinguipeng@whut.edu.cn; chudf@whut.edu.cn; luyuang@whut.edu.cn; wxg\_whut\_1701@whut.edu.cn).}
\thanks{$^{2}$Liping Lu is with the School of Computer Science and Artificial Intelligence, Wuhan University of Technology, Wuhan 430070, China (e-mail: luliping@whut.edu.cn).}
\thanks{$^{3}$Zejian Deng is with the Mechatronic Vehicle Systems Lab at the University of Waterloo,Waterloo, ON N2L3G1, Canada(e-mail: z49deng@uwaterloo.ca).}%

}

\providecommand{\keywords}[1]{\textbf{\textit{Index terms---}} #1}
\begin{document}

\maketitle

\begin{abstract}
Trajectory prediction is crucial for autonomous driving, as it aims to forecast the future movements of traffic participants.
Traditional methods usually perform holistic inference on the trajectories of agents, neglecting differences in prediction difficulty among agents.
This paper proposes a novel Difficulty-Guided Feature Enhancement Network (DGFNet), which leverages the prediction difficulty differences among agents for multi-agent trajectory prediction.
Firstly, we employ Spatio-temporal Feature Extraction to capture rich spatio-temporal features.
Secondly, a Difficulty-Guided Decoder controls the flow of future trajectories into subsequent modules, obtaining reliable future trajectories. 
Then, feature interaction and fusion are performed through the Future Feature Interaction module.
Finally, the fused actor features are fed into the Final Decoder to generate the predicted trajectory distributions for multiple participants.
Experimental results demonstrate that our model achieves SOTA performance on the Argoverse 1\&2 motion forecasting benchmarks. 
Ablation studies further validate the effectiveness of each module. 
Moreover, compared to the SOTA methods, our method balances trajectory prediction accuracy and real-time inference speed.
The code is available at \href{https://github.com/XinGP/DGFNet}{https://github.com/XinGP/DGFNet}.
\end{abstract}

\section{Introduction}
\label{sec:intro}

\IEEEPARstart{T}{rajectory} prediction is a crucial component of current autonomous driving systems. 
The objective of this task is to infer the future trajectory distribution of agents by considering their historical trajectories, dynamic and static scene information, and the interactions between these elements. 
By accounting for these complex factors, the future trajectory distribution can accurately reflect the agent's motion trends, ensuring the proper functioning of the autonomous driving system.

\begin{figure}[t]
	\centering
	\includegraphics[clip, trim=0.0cm 0cm 0.0cm 0.0cm, width=\linewidth]{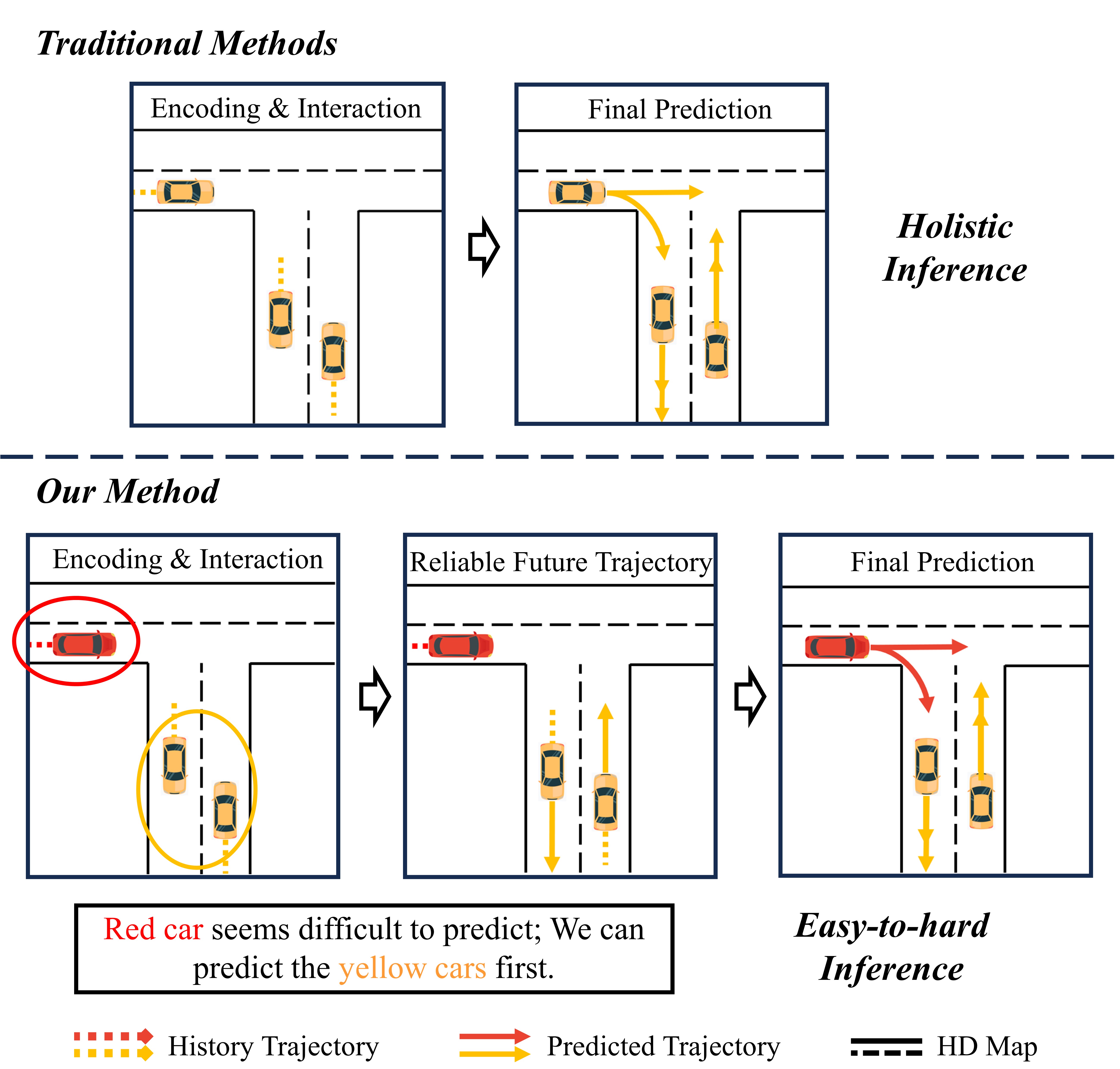}
        \vspace{-0.5cm}
        \setlength{\abovecaptionskip}{3pt}
	\caption{In contrast to traditional trajectory prediction methods, we have incorporated an intermediate step to obtain reliable future trajectories. The lower part of the figure illustrates the varying prediction difficulty among different vehicles in a sample traffic scenario. The future driving trajectory of the yellow vehicle is relatively easy to predict.}
	\label{fig:teaser}
        \vspace{-0.5cm}
\end{figure}

Previously, architectures of mainstream prediction models could be roughly divided into three steps: feature encoding, feature interaction, and feature decoding. 
Recently, some methods have improved the accuracy and robustness of prediction models through \textbf{feature enhancement}~\cite{zhao2021tnt,mangalam2020not,wang2023ganet,aydemir2023adapt,wang2023prophnet,kang2024ffinet,sun2022m2i,song2022learning,schmidt2023reset,park2023leveraging}. 
For example, some studies model the future intentions of vehicles as latent variables and enhance features based on these intentions~\cite{zhao2021tnt,mangalam2020not,wang2023ganet,aydemir2023adapt,park2023leveraging}.
Other studies model intermediate latent variables in the form of trajectories, where the outputs of the initial decoders are one or more predicted trajectories~\cite{wang2023prophnet,kang2024ffinet,sun2022m2i}.
In addition to network prediction methods, initial trajectories or intentions can also be obtained through planners or distance sampling~\cite{song2022learning,schmidt2023reset}.

Although these methods significantly improve model performance compared to mainstream model architectures, they neglect the inherent heterogeneity in prediction difficulty among agents, \textit{i.e.}, the natural differences in prediction difficulty for different agents within the scene. 
In real driving scenarios, human drivers often predict the future behaviors of most agents subconsciously, relying on intuition~\cite{gulati2021interleaving}. 
However, when faced with agents that may pose risks and exhibit high prediction difficulty, human drivers tend to carefully consider their behavior and motivation, such as those with tracking loss or low-quality trajectories due to perception issues~\cite{jiang2024social,li2023bcdiff,li2024itpnet}, or those exhibiting aggressive driving behaviors~\cite{schwarting2019social}.
As illustrated in the Fig.~\ref{fig:teaser}, The primary difference in our pipeline is the addition of an intermediate module that generates reliable future trajectories, thus better prediction results can be achieved for agents with higher prediction difficulty.

Moreover, due to the transition in trajectory prediction tasks from single-agent to multi-agent scenarios, the research trend in \textbf{scene representation} has shifted from scene-centric to agent-centric approaches.
However, embedding future trajectories \& anchors into agent-centric scene representation methods is highly inconvenient. 
This is because agent-centric scene representation methods require pairwise-relative pose to describe spatial positional relationships between lanes, agents, and other elements~\cite{jia2023hdgt,cui2023gorela,zhou2022hivt,zhang2024real,zhang2024simpl}. 
Consequently, agent-centric encoded features cannot comprehend the coordinate information of future trajectories and align features accordingly.
To address this, we aim to build a future feature enhancement network for multi-agent trajectory prediction by leveraging the strengths of both scene representation approaches.

Inspired by human drivers, we aim to investigate the effectiveness of prediction methods that follow the principle of easy-to-hard in multi-agent trajectory prediction tasks. The main contributions of this paper can be summarized as follows:

1) We introduce a novel network architecture that combines the strengths of two scene representations and leverages the differences in prediction difficulty among agents for multi-agent trajectory prediction, thereby improving overall prediction accuracy.

2) Inspired by human drivers, our method follows the principle of easy-to-hard. By initially predicting the easier-to-predict agents and using their reliable future trajectories for feature interaction and fusion, we enhance the prediction results for agents with higher prediction difficulty.

The organization of this paper is as follows: Section \ref{sec:meth} provides a detailed introduction to our pipeline; Section \ref{sec:exp} analyzes the experimental details and results of our proposed model and compares it with several SOTA methods; finally, Section \ref{sec:conclusion} summarizes our research and discusses future related research directions.

\section{Method}
\label{sec:meth}

\subsection{Overview}
\begin{figure*}[t]
	\centering
	\includegraphics[width=0.95\textwidth]{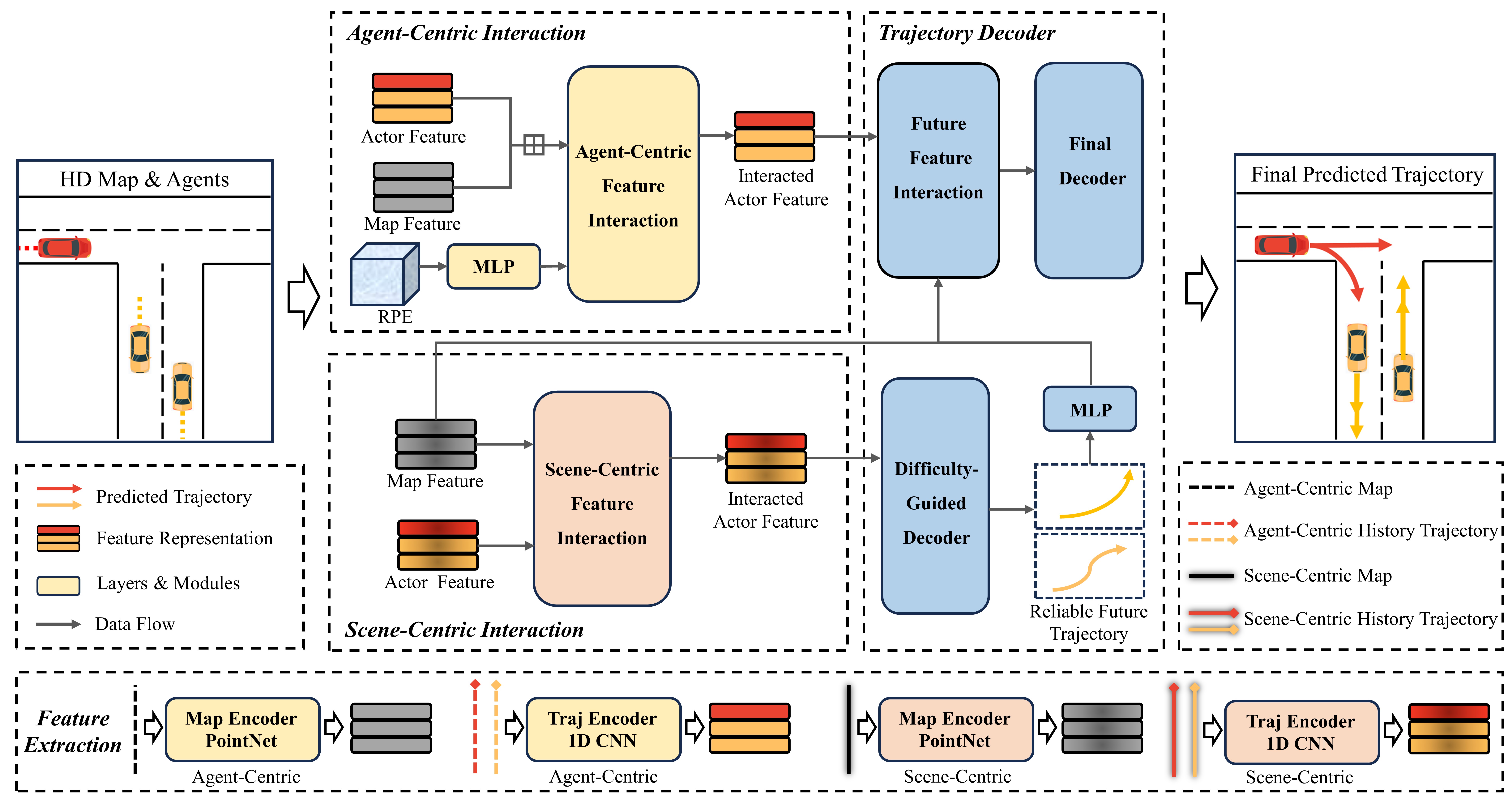}
    \vspace{-0.1cm}
    \setlength{\abovecaptionskip}{3pt}
	\caption{Our Spatio-temporal Feature Extraction includes two sets of independent encoders, which extract features based on scene representations (bottom). Subsequently, the extracted features pass through their respective Feature Interaction modules to obtain interacted actor features. Finally, we obtain the predicted trajectories and their corresponding probabilities through the trajectory decoder with Future Feature Enhancement and Difficulty-Guided Decoder.}
	\label{fig:method_full}
        \vspace{-0.5cm}
\end{figure*}

We propose DGFNet, a network designed for multi-agent trajectory prediction using a Difficulty-Guided Feature Enhancement module for asynchronous interaction between historical trajectories and reliable future trajectories.
As shown in Fig.~\ref{fig:method_full}, the architecture of DGFNet can be divided into three components: spatio-temporal feature encoding, feature interaction, and trajectory decoding.

\textbf{(1)Spatio-temporal Feature Extraction.} This component extracts spatio-temporal features separately from agent-centric historical trajectory information and scene-centric information using both agent-centric and scene-centric approaches.
\textbf{(2)Feature Interaction.} Unlike Spatio-temporal Feature Extraction, this component is not compatible with both scenarios.
However, feature interaction is not. The feature interaction component is categorized into two types: multi-head attention module and multi-head attention module with edge features.
The two interaction methods are respectively applicable to scene-centric and agent-centric scenario representations.
\textbf{(3)Trajectory Decoder.} This component is divided into a Difficulty-Guided Decoder and a Final Decoder. The Difficulty-Guided Decoder operates in the intermediate stage, while the Final Decoder generates the prediction results.

\begin{figure}[t]
	\centering
	\includegraphics[clip, trim=0.0cm 0cm 0.0cm 0.0cm, width=0.95\linewidth]{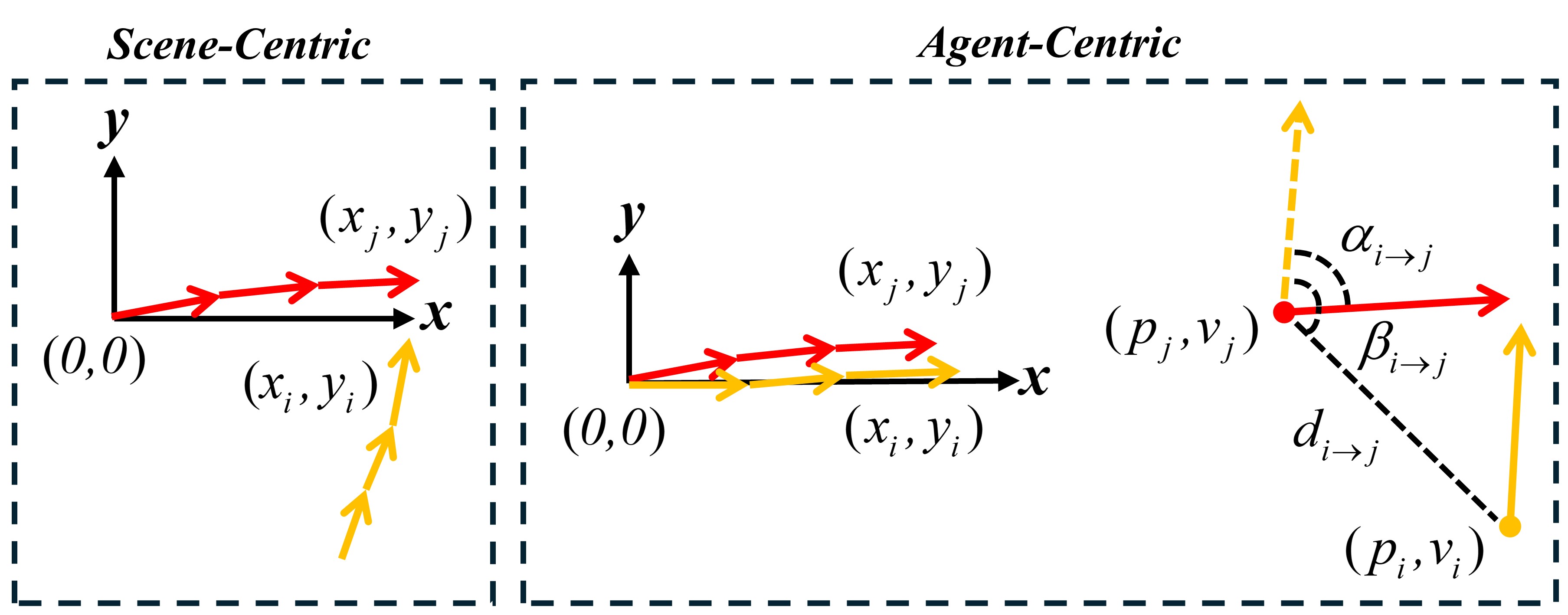}
        \vspace{-0.2cm}
        \setlength{\abovecaptionskip}{3pt}
	\caption{For the same scenario, the left side represents the scene-centric approach, which only requires using coordinate points. On the right side, the agent-centric approach necessitates expressing through local coordinate points and pairwise-relative poses.}
	\label{fig:rpe}
        \vspace{-0.5cm}
\end{figure}

\subsection{Problem Formulation}
Building on the majority of previous work, such as~\cite{Liang2020, Gao2020, zhao2021tnt, zhou2022hivt,schwarting2019social}, we assume that upstream perceptual tasks can provide high-quality 2D trajectory tracking data for agents in a coordinate system.
At the same time, localization and mapping tasks can provide accurate self-vehicle positioning data and comprehensive HD map data for the urban environment.
In other words, for $ j $ agents within a given scene, we obtain $ x $ and $ y $ positions, denoted as $ \mathcal{X}^{0:j} _{-t_{h} :0} $, corresponding to the time stamp horizon $\left \{ -t_{h},...,0,1,...,t_{f}  \right \} $.
The trajectory prediction task consists of predicting the future trajectories $ \mathcal{Y}^{0:j} _{1:t_{f}} $ of the agents by using the HD map information(including road coordinates, topological links, traffic lights) within the given scene and the historical trajectories output by the tracking task.
We represent the trajectory information of agents and lane information in vectorized form, where the agent trajectories include historical spatio-temporal sequences, and the map information includes lane coordinates, types, and traffic signal details.

It should be noted that this method involves two types of scenario representations. 
As shown in Fig.~\ref{fig:rpe}, the scene-centric representation involves only one coordinate system, where all agents and lane markings are expressed within this coordinate system. 
For simplicity, we only use uppercase notation to denote the trajectory and map tensor as $\mathcal{X}$ and $\mathcal{M}$, respectively.
In the agent-centric representation, multiple coordinate systems exist. 
Initially, all agents and lane markings are rotated using matrix rotation so that their orientations align with the $x$-axis. 
Subsequently, we use the original coordinates \(p\) and velocity orientation \(v\) to calculate the relative position distance \(d_{i\rightarrow j}\) and the angular difference \(\alpha_{i \rightarrow j}\), \(\beta_{i \rightarrow j}\).
We denote the agent-centric trajectory and map tensor as \( \mathcal{A} \) and \( \mathcal{L} \), respectively. Pairwise-relative pose is represented as \( \mathcal{RPE} = \left \{ ||d_{i \rightarrow j}||, \sin( \alpha_{i \rightarrow j} ), \cos( \alpha_{i \rightarrow j} ), \sin(\beta_{i \rightarrow j}),\cos(\beta_{i \rightarrow j})  \right \} \).

\subsection{Spatio-temporal Feature Extraction} 
In the problem formulation, we explain the process of representing agent and high-definition (HD) map data as vectors, establishing a corresponding mapping between continuous trajectories, map annotations, and sets of vectors. 
This vectorization method allows us to encode trajectory and map information as vectors. 
Following the trajectory feature extraction method of LaneGCN~\cite{Liang2020}, our trajectory feature extraction module mainly uses 1D CNN and downsampling techniques to encode the historical trajectories of all vehicles in the scene, thus obtaining encoded historical trajectory features.
For the map tensor, we use the PointNet~\cite{qi2017pointnet} to encode lane nodes and structural information to obtain map features.
It is worth mentioning that, when processing scene-centric vector information, we keep vehicle historical trajectories and lane line information separate.
Our view is that vehicle trajectories, as dynamic vectors, need to be distinguished from static maps to better capture the local motion characteristics of the vehicles.
The extraction of historical trajectory features is described as follows:
\begin{equation}
    \begin{split}
    \mathcal{Z}_i &= \rm {Conv1d}(\rm{Res1d}(\mathcal{X}_i)) , \quad i = 0, 1, \dots, n_{\text{fpn}} - 1, \\
    \hat{\mathcal{X}}_i &= \rm{Inter}(\mathcal{Z}_{i+1}, \rm{Scale}=2) + \mathcal{Z}_i \quad
    \label{eq:agent_encoder2}
    \end{split}
\end{equation}
after Res1d and Conv1d encoding, feature alignment is performed by scaling and interpolation operations.
For the agent-centric trajectory tensor, the same feature extraction is used to obtain  $\hat{\mathcal{A}}$.
The process of map feature extraction is described as follows:
\begin{equation}
    \begin{split}
    {\mathcal{H}}_1 &= \rm{PAB}_1\left(\rm{ReLU}(\rm{LayerNorm}(\mathcal{M} \textbf{W}_p + \textbf{b}_p))\right), \\
    \hat{\mathcal{M}} &= \rm{PAB}_2({\mathcal{H}}_1)
    \end{split}
\end{equation}
where \( \mathcal{M} \in \mathbb{R}^{N_{\text{lane}} \times 10 \times C_{\text{in}}} \) are the input lane features, \(\textbf{W}_p \in \mathbb{R}^{C_{\text{in}} \times h}\) and \(\textbf{b}_p \in \mathbb{R}^h\) are the weights and biases for the projection layer, and \(\rm{PAB}\) represents the PointAggregateBlock.
Similarly, we can obtain the agent-centric map feature $\hat{\mathcal{L}}$.
For $\mathcal{RPE}$ we use \(\rm{MLPs}\) for feature extraction to get $\hat{\mathcal{RPE}}$.

\subsection{Feature Interaction}
Our intention was to utilize the global representation of scene-centric scenes to embed future trajectory features. 
Interestingly, we found that even without any future enhancement, simply concatenating the actor features after the two interactions provides a noticeable improvement.
Specific results can be seen later in the ablation experiments.
Here we will present the two interaction methods: multi-head attention module and multi-head attention module with edge features, respectively.

We first introduce a generic feature interaction block.
In contrast to the general method, which uses simple attention operations for feature interaction, we believe that multi-head attention mechanisms are often better suited to handle complex input sequences and capture a wider range of hierarchical and diverse information.
We opt for Multi-Head Attention Blocks (MHAB) instead of the previously used simple attention operations, as used for the features~\cite{girgis2021latent}.
\begin{align}
    &\rm{MHA}(\textbf{Q}, \textbf{K}, \textbf{V}) = \rm{Softmax}\left(\frac{\textbf{Q} \textbf{K}^T}{\sqrt{dim_k}}\right) \textbf{V},
    \nonumber\\
    &\textbf{Q}, \textbf{K}, \textbf{V} = \textbf{W}^q, \textbf{W}^k, \textbf{W}^v
\end{align}
where $\rm{MHA}$ denotes the Multi-Head Attention operation.
$\textbf{Q}$, $\textbf{K}$, $\textbf{V}$ are computed by linear projections $\textbf{W}^q$, $\textbf{W}^k$, $\textbf{W}^v$ applied to input vectors.
The feature interaction module for carrying edge features is described as follows:
\begin{equation}
    \begin{split}
    &\rm{MHA}(\textbf{Q}', \textbf{K}', \textbf{V}') = \rm{Softmax}\left(\frac{\textbf{Q}' \textbf{K}'^T}{\sqrt{dim_k}}\right) \textbf{V}', \\
    &\textbf{Q}', \textbf{K}', \textbf{V}' = \textbf{Q} + \textbf{W}^q \text{edge}', \textbf{K} + \textbf{W}^k \text{edge}', \textbf{V} + \textbf{W}^v \text{edge}'
    \end{split}
\end{equation}
where \( \textbf{Q}', \textbf{K}', \textbf{V}' \) are the query, key, and value tensors after linear transformations that incorporate edge features, and \( \text{edge}' \) is the edge feature after linear transformation.
Finally, the normalized output after attention and dropout application is described as follows:
\begin{equation}
\hat{\textbf{x}} = \rm{LayerNorm}(\textbf{x} + \rm{Drop}(\rm{MHA}(\textbf{Q}', \textbf{K}', \textbf{V}')))
\end{equation}

For ease of expression, we denote regular multi-head attention as $\rm{MHA}$ and multi-head attention with edge features as $\rm{MHA}_{edge}$. 
Our Scene-centric Feature Interaction is described as follows:
\begin{equation}
\begin{split}
&\hat{\mathcal{X}}^{(l)} = \rm{MHA}_{MA}(\hat{\mathcal{M}}^{(l-1)}, \hat{\mathcal{X}}^{(l-1)}),\\
&\hat{\mathcal{M}}^{(l)} = \rm{MHA}_{MM}(\hat{\mathcal{M}}^{(l-1)}),\\
&\hat{\mathcal{M}}^{(l)} = \rm{MHA}_{AM}(\hat{\mathcal{X}}^{(l)}, \hat{\mathcal{M}}^{(l-1)}),\\
&\hat{\mathcal{X}}^{(l)} = \rm{MHA}_{AA}(\hat{\mathcal{X}}^{(l-1)})
\end{split}
\end{equation}
where \( l \) denotes the current layer index, and \(\hat{\mathcal{X}}^{(0)} = \hat{\mathcal{X}}\) and \(\hat{\mathcal{M}}^{(0)} = \hat{\mathcal{M}}\) are the initial inputs. 
Our Agent-centric Feature Interaction is described as follows:
\begin{equation}
\hat{\mathcal{A}}^{(k)} = \rm{MHA}_{edge}(\hat{\mathcal{A}}^{(k-1)}, \hat{\mathcal{L}}^{(k-1)}, \hat{\mathcal{RPE}}^{(k-1)})
\end{equation}
similarly, we complete multiple rounds of interactions by looping through multiple levels of updates.
\subsection{Trajectory Decoder}
As shown in Fig.~\ref{fig:method_full}, our Trajectory Decoder consists of three modules. First, the actor features from the scene-centric representation are passed into the Difficulty-Guided Decoder module to obtain reliable future trajectories and encode them. Then, through the Future Feature Enhancement module, the actor features from the agent-centric representation are expressed and aligned in the global coordinate system. Finally, the interacted features are passed into the Final Decoder to obtain the final prediction results.

\textbf{Difficulty-Guided decoder.}
To capture features of plausible future trajectories in the scene, we introduce a Difficulty-Guided Decoder to obtain the future trajectories of agents that are relatively easier to predict.
Firstly, we decode the actor features $\hat{\mathcal{X}}^{(l)}$ in scene-centric representation, which have undergone multiple layers of feature interactions. 
Each agent receives six predicted trajectories.
Following the ADAPT~\cite{aydemir2023adapt} trajectory decoder to get higher quality decoded trajectories by predicting endpoints for refinement.
The decoding process is described as follows:
\begin{equation}
\begin{split}
&\hat{\mathcal{E}} = \rm{MLP}_{end}(\hat{\mathcal{X}}^{(l)}) + \rm{MLP}_{refine}([\hat{\mathcal{X}}^{(l)}, \hat{\mathcal{E}}]),\\
&\hat{\mathcal{P}} = [\rm MLP_{traj}([\hat{\mathcal{X}}^{(l)}, \hat{\mathcal{E}}]), \hat{\mathcal{E}}]
\end{split}
\end{equation}
where endpoints \(\hat{\mathcal{E}}\) are predicted and refined by concatenating with the actor features \(\hat{\mathcal{X}}^{(l)}\), the trajectories \(\hat{\mathcal{P}}\) predicted with the refined endpoints and $[\cdot]$ represents the concatenation.

\begin{algorithm}[h]
\caption{Difficulty Masker for Predicted Trajectories} \label{algo:mask_process}
\begin{algorithmic}[1]
\renewcommand{\algorithmicrequire}{\textbf{Input:}}
\renewcommand{\algorithmicensure}{\textbf{Output:}}
\REQUIRE Predicted trajectories $\hat{\mathcal{P}}$, Predicted endpoints $\hat{\mathcal{E}}$, threshold $\tau$
\ENSURE Masked trajectories $\hat{\mathcal{P}}_{masked}$
\STATE $\hat{\mathcal{P}}_{masked} = set() $
\STATE $\bar{\mathcal{E}} \leftarrow \hat{\mathcal{E}}.mean()$ 
\FOR {$i = 1$ to $N$}
    \STATE $\mathcal{D} = \| \hat{\mathcal{E}} - \bar{\mathcal{E}} \|_2$
    \STATE $\bar{\mathcal{D}} \leftarrow \mathcal{D}.mean()$
    \IF {$\bar{\mathcal{D}} \leq \tau$}
        \STATE $\hat{\mathcal{P}}_{masked}$.add($\hat{\mathcal{P}}_i$)
    \ENDIF
\ENDFOR
\STATE
\RETURN $\hat{\mathcal{P}}_{masked}$
\end{algorithmic}
\end{algorithm}

\begin{table*}
    \centering
    \vspace{-0.2cm}
    \caption{Comparisons with the methods listed on the leaderboard of the Argoverse 1 motion Forecasting test set. The numbers highlighted in bold represent the best-performing results among them.}
    \renewcommand{\arraystretch}{1.1}
    \resizebox{0.9\textwidth}{!}{
    \begin{tabular}{c|c|cccccccc}
        \hline
\multirow{2}*{Inference} & \multirow{2}*{Methods} & \multirow{2}*{Year} &\hspace{-0.3cm} p-minFDE~$\downarrow$ \hspace{-0.3cm} & minFDE~$\downarrow$ &\hspace{0.1cm} MR~$\downarrow$ \hspace{0.1cm} & minADE~$\downarrow$& minFDE~$\downarrow$ & minADE~$\downarrow$ &\multirow{2}*{DAC~$\uparrow$} \hspace{-0.3cm} \\
~  & ~  & ~ & (K=6) & (K=6) & (K=6) & (K=6) & (K=1) & (K=1)\\
        \hline
        \multirow{9}{*}{Single model} & LaneGCN \cite{Liang2020} & 2020 & 2.05 & 1.36 & 0.162 & 0.87 & 3.76 & 1.70 & 0.9812\\
        & THOMAS \cite{gilles2021thomas}& 2021 & 1.97 & 1.44 & \textbf{0.104} & 0.94 & 3.59 & 1.67 & 0.9781\\
        & HiVT-128 \cite{zhou2022hivt}& 2022 & 1.84 & 1.17 & 0.127 & 0.77 & 3.53 & 1.60 & 0.9888\\
        & LAformer \cite{liu2024laformer}& 2023 & 1.84 & 1.16 & 0.125 & \underline{0.77} & \underline{3.45} & \underline{1.55} & 0.9897\\
        & HeteroGCN \cite{gao2023dynamic}& 2023 & 1.84 & 1.19 & 0.120 & 0.82 & 3.52 & 1.62 & -\\
        & Macformer \cite{feng2023macformer}& 2023 & 1.83 & 1.22 & 0.120 & 0.82 & 3.72 & 1.70 & \underline{0.9906}\\
        & GANet \cite{wang2023ganet}& 2023 & \underline{1.79} & \underline{1.16} & 0.118 & 0.81 & 3.46 & 1.59 & 0.9899\\
        & \textbf{DGFNet(single model)} & - & \textbf{1.74} & \textbf{1.11} & \underline{0.108} & \textbf{0.77} & \textbf{3.34} & \textbf{1.53} & \textbf{0.9909}\\
        \hline
        \multirow{9}{*}{Ensembled model} & HOME+GOHOME \cite{gilles2021home, gilles2022gohome}& 2022 & 1.86 & 1.29 & \textbf{0.085} & 0.89 & 3.68 & 1.70 & 0.9830\\
        & HeteroGCN-E \cite{gao2023dynamic}& 2023 & 1.75 & 1.16 & 0.117 & 0.79 & 3.41 & 1.57 & 0.9886\\
        & Macformer-E \cite{feng2023macformer}& 2023 & 1.77 & 1.21 & 0.127 & 0.81 & 3.61 & 1.66 & 0.9863\\
        & Wayformer \cite{nayakanti2023wayformer}& 2023 & 1.74 & 1.16 & 0.119 & 0.77 & 3.66 & 1.64 & 	0.9893\\
        & ProphNet \cite{wang2023prophnet}& 2023 & 1.69 & 1.13 & 0.110 & 0.76 & \textbf{3.26} & \textbf{1.49} & 0.9893\\
        & QCNet \cite{zhou2023query}& 2023 & \underline{1.69} & \textbf{1.07} & \underline{0.106} & \textbf{0.73} & 3.34 & \underline{1.52} & 0.9887\\
        & FFINet \cite{kang2024ffinet}& 2024 & 1.73 & 1.12 & 0.113 & 0.76 & 3.36 & 1.53 & 0.9875\\
        & \textbf{DGFNet(ensembled model)} & - & \textbf{1.69} & \underline{1.11} & 0.107 & \underline{0.75} & \underline{3.36} & 1.53 & \textbf{0.9902}\\ 
        \hline
    \end{tabular}}
    \vspace{-0.5cm}
    \label{tab: Argoverse leaderboard}
\end{table*}

The Argoverse 1 dataset shows that the predicted trajectories of most of the vehicles in the scenarios exhibit a high degree of concentration, with more than 90\% of the samples of straight ahead trajectories in each case~\cite{kim2022diverse}.
This suggests that the motion patterns of most agents can be easily captured, reflecting real-world traffic scenarios.
In order to prevent higher difficulty prediction trajectories from entering the subsequent modules and generating cumulative errors, we introduced a difficulty masker to filter the initial prediction results.
SceneTransformer\cite{ngiam2021scene}, inspired by recent approaches to language modeling, has pioneered the idea of using a masking strategy as a query to the model, enabling it to invoke a single model to predict agent behavior in multiple ways (marginal or joint prediction).
This mechanism in this paper is intended to mask out the initial prediction trajectories of the more difficult agents, thus providing better control over the prediction accuracy.
This process can be formalized as Algo.~\ref{algo:mask_process}.

\textbf{Future feature enhancement.}
For the reliable future trajectories, we first flatten them by performing the flatten operation on multiple future trajectories for feature dimension alignment. 
Then future trajectory feature extraction is performed by a simple MLP layer. 
These two operations can be described as follows:
\begin{equation}
\hat{\mathcal{F}} = \rm{MLP}(\rm{Flatten}(\hat{\mathcal{P}}_{masked}))
\end{equation}

Subsequently, we interact the future trajectory features $\hat{F}$ with the agent-centric interactive actor features $\hat{\mathcal{A}}^{(k)}$ via multi-head attention.
This operation aims to impose certain constraints on the prediction of other agents through the reliable future trajectory features.
Finally, we need to perform one last interaction using the original map features \(\hat{\mathcal{M}}\).
This step is crucial as it aims to refocus on the reachable lane lines of the map after imposing social constraints through future trajectory features\cite{dong2024proin}.
These two operations can be described as follows:
\begin{equation}
\begin{split}
&\mathcal{H} = \rm{MHA}_{AF}(\hat{\mathcal{A}}^{(k)}, \hat{\mathcal{F}}),\\
&\hat{\mathcal{O}} = \rm{MHA}_{AM}(\mathcal{H}, \hat{\mathcal{M}})
\end{split}
\end{equation}

\textbf{Final decoder.}
We perform feature fusion operations on the final features \(\hat{O}\) and the scene-centric feature \(\hat{X}^{(l)}\), expanding the decoding dimension to 256. 
Our final predictor can generate multimodal trajectories for all agents in the scene and their corresponding probabilities in a single inference. 
This process is described as follows:
\begin{equation}
\begin{split}
&{\mathcal{A}}_{fusion} = [\hat{\mathcal{O}}, \hat{\mathcal{X}}^{(l)}],\\
&\mathcal{E} = \rm{MLP}_{end}({\mathcal{A}}_{fusion}) + \rm{MLP}_{refine}([{\mathcal{A}}_{fusion}, \mathcal{E}]),\\
&\mathcal{K} = \rm{Softmax}(\rm{MLP}_{cls}([\mathcal{A}_{fusion}, \mathcal{E}])),\\
&\mathcal{P} = [\rm{MLP}_{traj}([{\mathcal{A}}_{fusion}, \mathcal{E}]), \mathcal{E}]
\end{split}
\end{equation}
where $\mathcal{P}$ represents the multimodal prediction trajectory, its corresponding probability $\mathcal{K}$ that we finally obtain and $[\cdot]$ represents the concatenation.

\textbf{Model training.} Although our feature enhancement component outputs trajectories during the intermediate process, our training process remains end-to-end.
Our model obtains initial and final predictions in a single pass.
Similar to the previous method~\cite{Liang2020, Gao2020, zhao2021tnt, zhou2022hivt}, we supervise the output trajectories $\mathcal{P}$ and $\hat{\mathcal{P}}$ during the training process by smoothing the $\mathcal{L}1$ loss, which is expressed as ${\mathcal{L}}_{r e g}$ and ${\mathcal{L}}_{r e g}^{c}$.
For the categorization loss in $\mathcal{K}$, we use the maximum marginal loss, which is expressed as ${\mathcal{L}}_{c l s}$.
Our overall loss expression is as follows:
\begin{equation}
\label{deqn_ex1a}
\mathcal{L}=\alpha{\mathcal{L}}_{r e g}+\beta{\mathcal{L}}_{c l s}+\lambda{\mathcal{L}}_{r e g}^{c}
\end{equation}
where $\alpha$, $\beta$ and $\lambda$ are three constant weights.

\section{Experiments}
\label{sec:exp}
\subsection{Experimental Setup}
\textbf{Datasets.}
Our model is tested and evaluated on a very challenging and widely used self-driving motion prediction dataset: the Argoverse 1\&2~\cite{chang2019argoverse,wilson2023argoverse}. 
Both motion prediction datasets provide agent tracking trajectories and semantically rich map information at a frequency of 10Hz over a specified time interval. 
The prediction task in Argoverse 1 is to predict trajectories for the next 3 seconds based on the previous 2 seconds of historical data. The dataset contains a total of 324,557 vehicle trajectories of interest extracted from over 1000 hours of driving. 
Argoverse 2 contains 250,000 of the most challenging scenarios officially filtered from the self-driving test fleet.
Argoverse 2 predicts the next 6 seconds from the first 5 seconds of historical trajectory data.
To ensure fair comparisons between models, both datasets were subjected to official data partitioning and the test set was evaluated using the Eval online test server.

\textbf{Evaluation metrics.}
We have adopted the standard testing and evaluation methodology used in motion prediction competitions to assess prediction performance.
Key metrics for individual agents include Probabilistic minimum Final Displacement Error  (p-minFDE), Minimum Final Displacement Error (minFDE), Minimum Average Displacement Error (minADE), Miss Rate (MR) and Drivable Area Compliance (DAC).
Where p-minFDE, MR, and minFDE reflect the accuracy of the predicted endpoints, and minADE indicates the overall bias in the predicted trajectories. DAC reflects the compliance of the predicted outcomes.
The key metrics for multiple agents are obtained by averaging the performance metrics of the labeled agents in each scenario of the test dataset. 
It is worth noting that these labeled agents share the same scene prediction probability.

\textbf{Implementation details.}
We generate lane vectors for lanes that are more than 50 meters away from any available agent.
The number of layers $l$ and $k$ of interaction features are both set to 3.
All layers except the Final Decoder have 128 output feature channels. In addition, the weight parameters in the loss function are set to $\alpha$=0.7, $\beta$=0.1 and $\lambda$=0.2.
The hyperparameter $\tau$ in the difficulty masker was set to 5.
The model was trained 50 epochs on an Nvidia RTX 3090 with a batch size of 16. For the experiments on Argoverse 1\&2, we used a segmented constant learning rate strategy: up to the 5th epoch we used \(5 \times 10^{-5}\); from the 6th epoch to the 40th, we used \(5 \times 10^{-4}\); and thereafter, we used \(5 \times 10^{-5}\) until training was complete.

\subsection{Quantitative Results}
Quantitative evaluations on the Argoverse 1\&2 datasets demonstrate that DGFNet exhibits significant performance advantages across various metrics compared to SOTA methods. 
Our method remains competitive even when compared to ensemble strategies used by other top models like QCNet\cite{zhou2023query} and ProphNet\cite{wang2023prophnet}. 
Through ablation experiments, we demonstrate the effectiveness of each module mechanism and the experiments on adjusting the relevant parameters.
Computational efficiency analysis shows that DGFNet achieves high prediction accuracy with relatively fewer parameters, ensuring its practical applicability in real-time autonomous driving scenarios.

\begin{table}[h]
    \centering
    \vspace{-0.2cm}
     \caption{Comparisons with the methods listed on the leaderboard of the Argoverse 2 single-agent Forecasting.}
     \renewcommand{\arraystretch}{1.1}
     \resizebox{0.47\textwidth}{!}{
    \begin{tabular}{c|ccccc}
    \hline
         \multirow{2}*{Method} & \multirow{2}*{Year} & p-minFDE~$\downarrow$ & minADE~$\downarrow$ & minFDE~$\downarrow$ & MR~$\downarrow$ \\
         ~  & ~  & (K=6) & (K=6) & (K=6) & (K=6) \\
        \hline
        LaneGCN\cite{Liang2020} & 2020 & 2.64 & 0.91  & 1.96 & 0.30 \\
        THOMAS\cite{gilles2021thomas} & 2021 & 2.16 & 0.88  & 1.51 & 0.20 \\
        HDGT\cite{jia2023hdgt} & 2023 & 2.24 & 0.84  & 1.60 & 0.21 \\
        HPTR\cite{zhang2024real} & 2023 & 2.03 & 0.73  & 1.43 & 0.19 \\
        GoReal\cite{cui2023gorela} & 2022 & 2.01 & 0.76  & 1.48 & 0.22 \\
        GANet\cite{wang2023ganet} & 2023 & 1.96 & 0.72  & \textbf{1.34} & \underline{0.17} \\
        \textbf{DGFNet} & - & \textbf{1.94} & \textbf{0.70} & 1.35  & \textbf{0.17}  \\
        \hline
    \end{tabular}}
    \label{tab:av2.1}
    \vspace{-0.2cm}
\end{table}

\begin{table}[h]
    \centering
    \vspace{-0.2cm}
     \caption{Comparisons with the methods listed on the leaderboard of the Argoverse 2 muti-world Forecasting.}
     \renewcommand{\arraystretch}{1.1}
     \resizebox{0.49\textwidth}{!}{
     {\Large
    \begin{tabular}{c|ccccc}
    \hline
         \multirow{2}*{Method} & \multirow{2}*{Year} & avgP-MinFDE~$\downarrow$ & avgMinADE~$\downarrow$ & avgMinFDE~$\downarrow$ & actorMR~$\downarrow$ \\
         ~  & ~  & (K=6) & (K=6) & (K=6) & (K=6) \\
        \hline
        LaneGCN\cite{Liang2020} & 2020 & 3.90 & 1.49  & 3.24 & 0.37 \\
        HiVT\cite{zhou2022hivt} & 2022 & 2.85 & 0.88  & 2.20 & 0.26 \\
        FJMP\cite{rowe2023fjmp} & 2023 & 2.59 & 0.81  & 1.89 & 0.23 \\
        FFINet\cite{kang2024ffinet} & 2024 & 2.44 & 0.77  & 1.77 & 0.24 \\
        \textbf{DGFNet} & - & \textbf{2.37} & \textbf{0.73} &  \textbf{1.68} & \textbf{0.20}  \\
        \hline
    \end{tabular}}}
    \label{tab:av2.2}
    \vspace{-0.2cm}
\end{table}

\textbf{Comparison with prediction performance in Agoverse 1.}
We conducted performance comparisons on the Argoverse 1 test set using both single-model and multi-model approaches between our method and the most classical and latest methods.
Generally, ensemble methods involve using K-means to match the closest trajectories from multiple models, followed by weighted computation to obtain the final result.
We obtained 5 models by setting different initial random seeds and performed ensemble inference using the aforementioned method. 
The results are shown in Tab.\ref{tab: Argoverse leaderboard}. Compared to single-model methods, our approach consistently ranks at the top in almost all metrics. 
Even when compared with methods using ensemble strategies, our approach remains highly competitive, achieving prediction accuracy comparable to the leading models such as QCNet\cite{zhou2023query} and ProphNet\cite{wang2023prophnet} in the leaderboard. 
However, our method has advantages in terms of parameter size and inference time, which will be detailed in the following sections.

\textbf{Comparison with prediction performance in Agoverse 2.}
As shown in Tab.\ref{tab:av2.1} \& Tab.\ref{tab:av2.2} , in our comparison of several trajectory prediction methods on the Argoverse 2 single-agent and multi-agent test sets, we found that while the advantage of our method is not as pronounced as in Argoverse 1, it remains reasonably acceptable. 
We speculate that this is primarily due to the increased prediction horizon in the Argoverse 2 dataset, which has led to less robust initial prediction results. 
Moreover, due to the inclusion of categories such as pedestrians and cyclists in Argoverse 2, the effectiveness of the parameter settings for our difficulty masker has also been impacted.
Overall, our model demonstrates highly competitive performance within an acceptable range of model parameters across both datasets.

\begin{table}[h]
    \Large
    \centering
    \caption{Comparison of prediction performance (minFDE~$\downarrow$) on high-difficulty samples in the Argoverse 1 validation set.}
    \renewcommand{\arraystretch}{1.2}
    \resizebox{0.48\textwidth}{!}{
    \begin{tabular}{c|c|c|c|c|c|c}
    \hline
         Method & Top 1\% & Top 2\% & Top 3\% & Top 4\% & Top 5\% & ALL \\
         \hline
         
        LaneGCN & 11.52 & 8.24 & 6.98 & 6.67 & 5.65 & 1.08 \\ 

        Agent-Base & 9.12\text{↓21\%} & 6.86\text{↓17\%} & 5.74\text{↓18\%} & 5.05\text{↓24\%} & 4.57\text{↓19\%} & 0.94\text{↓13\%} \\ 
        
        \textbf{DGFNet} & 7.26\textbf{↓37\%} & 5.21\textbf{↓36\%} & 4.36\textbf{↓37\%} & 3.85\textbf{↓42\%} & 3.51\textbf{↓30\%} & 0.89\textbf{↓17\%} \\

        \hline
    \end{tabular}}
    \label{tab:cw}
    \vspace{-0.2cm}
\end{table}

\textbf{Comparison of prediction performance on high-difficulty samples.}
Tab.\ref{tab:cw} shows the highest minFDE errors of each method on the Argoverse 1 validation set. 
We reasoned on the validation set with LaneGCN's\cite{Liang2020} pre-trained model to obtain some sample sets with the highest errors and then reasoned on these sample sets using our Agent-centric base model and DGFNet. 
It can be observed that the average accuracy difference for all samples is minimal compared to our base model. 
However, our method has significantly lower minFDE errors across all percentage ranges, clearly demonstrating the remarkable advantage of our method in mitigating the long-tail problem in trajectory prediction.

\textbf{Quantitative results and visualization.}
Fig.\ref{fig:vis} shows the qualitative visualization.
The four selected scenarios have high prediction difficulty and strong representativeness.
In Seq.1 and Seq.2, the model generates accurate predictions for each agent in a busy and strongly interacting intersection scenario.
In Seq.3, the model gives robust predictions in the face of vehicle turning tendencies.
As a comparison, in Seq.4 the model gives more scattered prediction results when the intention of the vehicle is not clear.
Overall, the model captured the final landing point very accurately in both turning and straight ahead situations.

\begin{table}[h]
    \centering
    \vspace{-0.2cm}
    \caption{The ablation study results of different components on the Argoverse 1 validation set.}
    \renewcommand{\arraystretch}{1.1}
    \resizebox{0.45\textwidth}{!}{
    \begin{tabular}{c|c|c|c|ccc}
    \hline
         \multirow{2}*{Base Model} & \multirow{2}*{DM} & \multirow{2}*{FFI}& \multirow{2}*{SFF}  & p-minFDE~$\downarrow$ & minADE~$\downarrow$ & minFDE~$\downarrow$ \\
         ~  & ~  & ~  & ~ & (K=6) & (K=6) & (K=6)  \\
         \hline
         Scene-Base & $\times$ & $\times$ &$\times$ & 1.641 & 0.699 & 1.044 \\
         Agent-Base & $\times$ & $\times$ &$\times$ & 1.559 & 0.657 & 0.941 \\
         - & $\times$ & $\times$ &\checkmark &1.525 & 0.646 & 0.916 \\
         - & $\times$ & \checkmark & \checkmark & 1.654 & 0.706 & 1.052
          \\
          - & \checkmark & $\times$ &\checkmark &1.550 & 0.652 & 0.934
          \\
        \textbf{DGFNet} & \checkmark & \checkmark & \checkmark & \textbf{1.492}& \textbf{0.632} & \textbf{0.892} \\
        \hline
    \end{tabular}}
    \label{tab:cp}
    \vspace{-0.1cm}
\end{table}

\begin{table}[h]
    \centering
    \vspace{-0.2cm}
    \caption{Performance versus loss fuction weight parameter of DGFNet evaluated on the Argoverse 1 validation set.}
    \renewcommand{\arraystretch}{1.1}
    \resizebox{0.30\textwidth}{!}{
    \begin{tabular}{cc|ccc}
    \hline
         \multirow{2}*{$\alpha$} & \multirow{2}*{$\lambda$} & p-minFDE~$\downarrow$ & minADE~$\downarrow$ & minFDE~$\downarrow$ \\
         ~   & ~   & (K=6) & (K=6) & (K=6)  \\
         \hline
         
        0.5 & 0.4 & 1.519 & 0.640 & 0.911 \\

        0.6 & 0.3 & 1.502 & 0.635 & 0.900 \\

        0.7 & 0.2 & \textbf{1.492} & \textbf{0.632} & \textbf{0.892} \\

        0.8 & 0.1 & 1.503 & 0.636 & 0.902 \\

        \hline
    \end{tabular}}
    \label{tab:los}
    \vspace{-0.6cm}
\end{table}

\begin{table}[h]
    \centering
    \vspace{-0.2cm}
    \caption{Performance versus $\tau$ parameter of DGFNet evaluated on the Argoverse 1 validation set.}
    \renewcommand{\arraystretch}{1.1}
    \resizebox{0.26\textwidth}{!}{
    \begin{tabular}{c|ccc}
    \hline
         \multirow{2}*{$\tau$} & p-minFDE~$\downarrow$ & minADE~$\downarrow$ & minFDE~$\downarrow$ \\
         ~   & (K=6) & (K=6) & (K=6)  \\
         \hline
         
        7 & 1.510 & 0.642 & 0.907 \\

        5 & \textbf{1.492} & \textbf{0.632} & \textbf{0.892} \\
        
        3 & 1.499 & 0.634 & 0.897 \\

        1 & 1.508& 0.639 & 0.905 \\
        
        \hline
    \end{tabular}}
    \label{tab:cp1}
    \vspace{-0.7cm}
\end{table}

\subsection{Analysis and Discussion}
\textbf{Ablation study.}
We conducted an ablation study on the Argoverse 1 validation set. 
As shown in \ref{tab:cp}, our base model only utilizes agent-centric or scene-centric historical trajectories and maps for feature extraction and interaction.
The third model performs Scene Feature Fusion (SFF) on agent-centric and scene-centric features, achieving a certain level of performance improvement.
The fourth model incorporates the Future Feature Interaction (FFI) module, enabling interaction and alignment between the extracted historical features, original map features, and future features.
The fifth model removes the SFF module and introduces the Difficulty Masker (DM), where the filtered future trajectory features are directly embedded without undergoing the interaction process.
\begin{equation}
\begin{split}
&\rm{SceneBase} = \rm{FiDecoder}(\hat{\mathcal{X}}^{(l)}),\\
&\rm{AgentBase} = \rm{FiDecoder}(\hat{\mathcal{A}}^{(k)}),\\
&\rm{w/\:SFF} = \rm{FiDecoder}([\hat{\mathcal{A}}^{(k)}, \hat{\mathcal{X}}^{(l)}]),\\
&\rm{w/\:SFF\&DM} = \rm{FiDecoder}([\hat{\mathcal{A}}^{(k)}, \hat{\mathcal{F}}]),\\
\end{split}
\end{equation}
where $\rm{SceneBase}$ represents the prediction results obtained by our scene-centric base model, $\rm{AgentBase}$ represents the prediction results obtained by our agnet-centric base model, $\rm{FiDecoder}$ represents the entire Final Decoder module,  $\rm{w/\:SFF}$ and $\rm{w/\:SFF\&DM}$ represent the prediction results of the base model with $\rm{SFF}$ and $\rm{SFF\&DM}$ added, respectively. $[\cdot]$ represents the concatenation.

Of concern is that the performance of the fifth model drops significantly, because the second stage model may find shortcuts during optimization if the results given in the first stage are good enough~\cite{geirhos2020shortcut}.

Regarding the selection of loss function weights, based on our previous experience in adjusting parameters on the base model and the principle that the most important weight should take the largest proportion, we first fixed the weight of \({\mathcal{L}}_{cls}\) (\(\beta\)) to 0.1. Then, we conducted ablation experiments by gradually increasing the weight of \({\mathcal{L}}_{reg}\) (\(\alpha\)) starting from 0.5. As shown in \ref{tab:los}, we found that the model performance was quite good when \(\alpha = 0.7\), \(\beta = 0.1\), and \(\lambda = 0.2\).

We also did ablation experiments on the effect of parameter $\tau$ on the model.
According to Tab.\ref{tab:cp1}, the model performance obtained by setting parameter $\tau$ to 5 is the best.

\begin{figure}[t]
	\centering
	\includegraphics[clip, trim=0.0cm 0cm 0.0cm 0.0cm, width=0.9\linewidth]{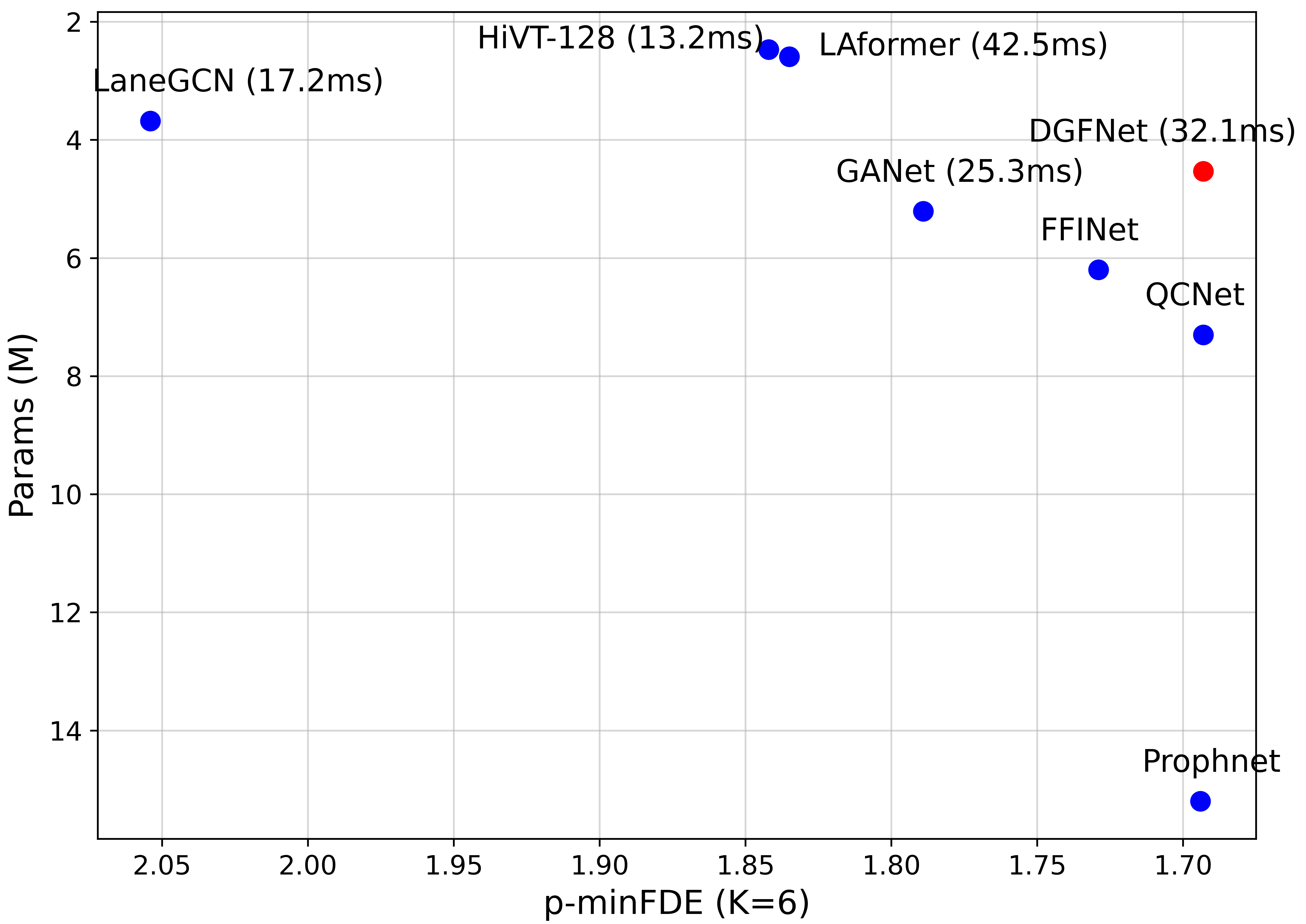}
        \vspace{-0.2cm}
        \setlength{\abovecaptionskip}{3pt}
	\caption{The x-axis represents the predicted result p-minFDE, and the y-axis represents the model's parameter size. Both values should be minimized for better performance.}
	\label{fig:ccp}
        \vspace{-0.5cm}
\end{figure}
\textbf{Computational performance.}
We compared the parameters of each method and the average inference time per scenario on the same device. 
As shown in Fig.\ref{fig:ccp}, our inference time is slightly higher than LaneGCN\cite{Liang2020} and close to the lighter models HiVT-128\cite{zhou2022hivt} and LAformer\cite{liu2024laformer}. 
However, our method has a much higher prediction accuracy than these two methods. 
Compared with the methods with the highest prediction accuracy, our method has fewer parameters, which means that our model achieves high prediction accuracy with lower computational cost. 
Then, we recorded the inference time on the same experimental machine equipped with an NVIDIA GeForce RTX 3090. 
The model inference time indicates that the real-time latency of DGFNet is within an acceptable range. 
Since LAformer uses a two-stage inference approach, its inference time is longer despite having fewer parameters than our method.

\section{Conclusion}
\label{sec:conclusion}
In this paper, we propose a novel \textbf{D}ifficulty-\textbf{G}uided \textbf{F}eature Enhancement Network (DGFNet) for muti-agent trajectory prediction.
Distinguishing from general future enhancement networks, our model emphasizes filtering future trajectories by masking out and retaining only reliable future trajectories for feature enhancement, which greatly improves the prediction performance. 
Extensive experiments on the Argoverse 1\&2 benchmarks show that our method outperforms most SOTA methods in terms of prediction accuracy and real-time processing. 
Emulating the predictive habits of human drivers is an intriguing direction for future research.
This involves emulating human strategies when encountering different vehicles and making effective assumptions in the presence of incomplete perceptual information.

\begin{figure*}[!]
\vspace{-0.5cm}
	\centering
	\includegraphics[width=0.95\textwidth]{at2ral.jpg}
    \vspace{-0.2cm}
	\caption{Quantitative results of DGFNet on the Argoverse 1\&2 validation set. The top scenarios correspond to the Argoverse 1 dataset and the bottom scenarios correspond to the Argoverse 2 dataset.}
	\label{fig:vis}
\vspace{-0.5cm}
\end{figure*}









\bibliography{references}

\end{document}